\documentclass[letterpaper, 10 pt, conference]{ieeeconf}  
\usepackage{commenting}
\usepackage{listings}
\usepackage{booktabs}
\usepackage{multirow}
\usepackage{todonotes}
\usepackage{wrapfig}
\usepackage[most]{tcolorbox}
\definecolor{cocoabrown}{rgb}{0.82, 0.41, 0.12}
\usepackage{tikz}
\usepackage{changepage}
\definecolor{ao(english)}{rgb}{0.0, 0.5, 0.0}
\definecolor{burntsienna}{rgb}{0.91, 0.45, 0.32}
\definecolor{bleudefrance}{rgb}{0, 0.5, 1}
\definecolor{etonblue}{rgb}{0, 0.8, 0}
\usepackage{varwidth}
\usepackage{tabularx}

\usepackage{booktabs}
\usepackage{tikz}
\usepackage{graphicx}
\usepackage{subfigure}
\usepackage{dblfloatfix}

\IEEEoverridecommandlockouts                              

\title{\LARGE \bf
Large Language Models for Orchestrating Bimanual Robots 
}
 
\author{
{Kun Chu}$^{*}$, {Xufeng Zhao}, {Cornelius Weber}, {Mengdi Li}, {Wenhao Lu}, and {Stefan Wermter}
\thanks{This research was funded by the German Research Foundation (DFG) in the project Crossmodal Learning (TRR-169) and by the China Scholarship Council (CSC).}
\thanks{The authors are with the Knowledge Technology Group, Department of Informatics, University of Hamburg, 22527 Hamburg, Germany. E-mail:
        {\tt\small {\{kun.chu, xufeng.zhao, cornelius.weber, wenhao.lu, stefan.wermter\}@uni-hamburg.de, mengdi.li@studium.uni-hamburg.de.}}}
\thanks{$^{*}$Corresponding author.}
}

\begin{document}

\maketitle
\thispagestyle{empty}
\pagestyle{empty}

\begin{abstract}
Although there has been rapid progress in endowing robots with the ability to solve complex manipulation tasks, generating control policies for bimanual robots to solve tasks involving two hands is still challenging because of the difficulties in effective temporal and spatial coordination. With emergent abilities in terms of step-by-step reasoning and in-context learning, Large Language Models (LLMs) have demonstrated promising potential in a variety of robotic tasks. However, the nature of language communication via a single sequence of discrete symbols makes LLM-based coordination in continuous space a particular challenge for bimanual tasks. To tackle this challenge, we present \textbf{LA}nguage-model-based \textbf{B}imanual \textbf{OR}chestration (LABOR), an agent utilizing an LLM to analyze task configurations and devise coordination control policies for addressing long-horizon bimanual tasks. We evaluate our method through simulated experiments involving two classes of long-horizon tasks using the NICOL humanoid robot. Our results demonstrate that our method outperforms the baseline in terms of success rate. Additionally, we thoroughly analyze failure cases, offering insights into LLM-based approaches in bimanual robotic control and revealing future research trends. The project website can be found at \url{http://labor-agent.github.io}.

\end{abstract}

\section{Introduction}
While humans routinely operate with both hands in their daily lives, achieving effective bimanual coordination remains a formidable challenge in robotics as the dual-arm setting faces complexity in terms of temporal and spatial coordination. Bimanual manipulation offers a significant difficulty for the existing planning-based and learning-based methods. Previous planning-based methods \cite{Vahrenkamp2011, Salehian2017, Ng2023} largely focus on motion planning for controlling two arms in manipulating large objects under certain constraints, limiting their application in dynamic or complex task spaces. On the other hand, learning-based methods, such as Reinforcement Learning (RL) and Imitation Learning (IL), enable a robot to learn control policies from human-designed rewards \cite{Luck2017, Chitnis2020} or human-teleoperated demonstrations \cite{Zollner2004, Stepputtis2022}. However, it is known that RL algorithms are notorious as hard to train, especially in dual-arm settings that have high degrees of freedom (DoFs), and collecting human demonstrations for bimanual robots is labor-intensive and time-consuming. At the same time, the high complexity associated with the variety of bimanual patterns suggests that low-level control policies and high-level planning must be considered for an integrated control system design \cite{Smith2012}.

\begin{figure}[t!]
\centering
\includegraphics[width=1\linewidth]{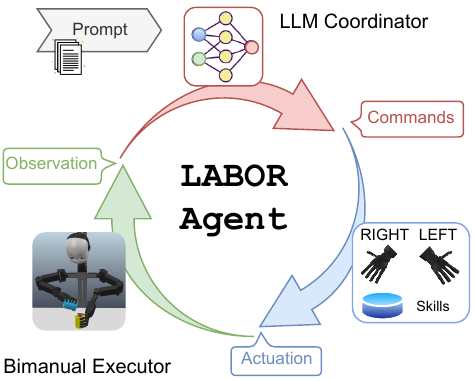}
\caption{Illustration of the LABOR agent. During the execution of the task, with the guiding prompt, the LLM coordinator first decomposes the task and then generates the step action plan, including the control command for the left and right hand. The bimanual robot executor then performs actions on the environment according to the commands, and the results provide feedback to the LLM for the next action, and so on until the end of the task.}
\label{fig:LABOR}
\end{figure}

Large Language Models (LLMs) have demonstrated remarkable knowledge and reasoning capabilities \cite{Wei22ChainofThoughtPrompting, Zhao24EnhancingZeroshot}, leading to a revolutionary change in numerous fields \cite{Bubeck23SparksArtificial}. In robotics, an LLM is typically employed as a high-level planner to reason and invoke primitive skills for accomplishing complex tasks described in natural language \cite{Ahn2022, Zhao23ChatEnvironment, Zhang23BootstrapYour}. 
However, existing methods primarily focus on uni-manual manipulations with complex chains of skills and 
the integration of LLMs in bimanual control remains unexplored.
In this research, we explore further the feasibility of leveraging LLMs in robotic bimanual control, especially focusing on high-level temporal and spatial coordination between two hands. Our research question is: \textit{How can LLMs be employed to achieve versatile bimanual manipulations in tasks with complex spatial relationships?}

To this aim, we introduce the \textbf{LABOR} (\textbf{LA}nguage-model-based \textbf{B}imanual \textbf{OR}chestration) agent, which uses an LLM to effectively coordinate bimanual control to deal with complex manipulation tasks, as shown in Fig. \ref{fig:LABOR}. Specifically, we introduce spatio-temporal bimanual control patterns and design the guiding prompt for LLMs. The prompted LLM decomposes the task execution into corresponding uncoordinated and coordinated stages to account for the spatial relationships and task requirements. Then it generates the step action commands for two hands respectively based on a pre-defined skill library. This proceeds iteratively until the long-horizon task is completed. We evaluate the LABOR agent on a simulation of the semi-humanoid bimanual NICOL, the Neuro-Inspired COLlaborator \cite{Matthias2023}, on two classes of challenging long-horizon tasks. Compared to the Baseline agent, the higher success rate of the LABOR agent on these tasks indicates its efficacy in coordinating both hands to perform complex long-horizon tasks.

\section{Related Work}
In this section, we describe the related work on planning-based and learning-based methods for bimanual manipulations, and LLM's use in robotics.

\subsection{Bimanual Manipulation}

\subsubsection{Planning-based methods}
Early works \cite{Koga1992, Vahrenkamp2011, Smith2012, Nemec2016} mainly focus on planning algorithms for dual-arm settings under the closed-chain constraint, which rely on static known dynamic models to generate kinematic configurations for two arms. With engineered primitives or pre-defined trajectories \cite{Salehian2017, Lertkultanon2018, Losey2020, Ng2023}, bimanual robots exhibit some flexibility in dealing with complex environments. However, such planning-based methods mainly rely on task-specific primitives, which are costly to design or collect in a more diverse task setting. 

\subsubsection{Learning-based methods}
Some works have employed RL \cite{Chen2022, Chitnis2020, Luck2017, Amadio2019} and IL \cite{Zollner2004, Lioutikov2016, Ureche2018, Stepputtis2022} framework in dual-arm settings. These approaches typically require massive samples and training time, due to the high dimensionality of the action space in a bimanual setting. Although some recent works proposed to design parameterized primitives to reduce the size of the action space for exploration, these movement primitives normally require costly hard-coded motion design or human-teleoperated demonstration collection, limiting their applications in complex environments. 

The approaches listed above do little to consider the properties of the task at a higher level but rather adopt a single control policy for controlling the robot to accomplish the task. In this sense, the control policy in their tasks is normally fixed and monolithic which tackles only simple and short-duration tasks. In contrast, we explore the usage of LLMs to generate control policies from a higher perspective that considers the uncoordinated and coordinated processes for task accomplishment, thus adapting to complex long-horizon tasks.

\subsection{LLM in Robotics}
Learning robotic control policies to address various manipulation tasks often requires the collection of massive datasets \cite{Kroemer2021, Li*23InternallyRewarded, Zhao22ImpactMakes, Yang23FoundationModels}. Recent works have leveraged common sense and established knowledge from LLMs to control robots in tackling complex tasks. 
Some studies have delved into distilling LLM expertise into localized models. For instance, researchers have utilized LLMs' programming capabilities to generate executable code for fulfilling open instructions \cite{Huang23VoxPoserComposable, Wang23VoyagerOpenEnded, Liang22CodePolicies}, or to provide direct guidance for RL policies \cite{Ma23EurekaHumanLevel, Chu23AcceleratingReinforcement}. Other approaches involve decomposing complex tasks into manageable sub-tasks using LLMs, thus enabling robots to address them individually with pre-defined skills \cite{Zhao23ChatEnvironment, Ahn2022}, i.e., language-conditioned policies.
We adopt the latter paradigm to construct a versatile robot equipped with atomic skills that can be efficiently reused across multiple tasks or potentially enhanced for new abilities \cite{Zhang23BootstrapYour}. Specifically, our robot is furnished with general skills for both arms, offering a vast number of potential combinations. However, this inherent complexity of combinatorial capabilities poses challenges to efficient reasoning and planning in bimanual manipulation, which will be explored in this study.


\section{Method}
This section introduces the LABOR agent from two perspectives: 
1) Spatio-temporal Bimanual Control, where we present a series of control patterns that form the foundation for the agent's practical design;
2) LLM Orchestration, where we describe the design of an LLM-based pipeline for implementing various control patterns in bimanual control.

Bimanual manipulation control mainly involves the consideration of coordination and, when required, synchronization. In a macroscopic view of long-horizon bimanual manipulation tasks, a combination of coordination stages can be involved. 
For example, in the task of placing an apple in a bowl and holding the bowl to a serving position, 
both uncoordinated processes (grasping and releasing the apple) and symmetrically coordinated processes (holding and moving the bowl with two hands) are required. 
During task execution, the type of control patterns should first be decided based on the current spatial and temporal situations, and then specific control commands should be generated. 

\begin{figure}[htp]
\centering
\includegraphics[width=1\linewidth]{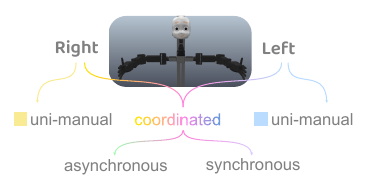}
\caption{Spatio-temporal control adopted by the LABOR agent.}
\label{fig:bimanual_control}
\end{figure}
\subsection{Spatio-temporal Bimanual Control}
Inspired by earlier works \cite{Zollner2004, Boehm2021, Krebs2022} on dual-arm manipulations, as shown in Fig. \ref{fig:bimanual_control}, we introduce the following spatio-temporal control patterns for two hands of a bimanual robot to manipulate on a worktable as follows,

\begin{itemize}
\item Uncoordinated Control: two hands act and manipulate independently in their own working spaces (i.e., \textit{left area} and \textit{right area} respectively), referred to as uni-manual manipulation for each hand. For safety concerns and better efficiency, the \textit{left area} is not accessible by the right hand, and vice versa. 

\item Coordinated Control: two hands act and manipulate dependently in terms of either spatial or temporal relation, e.g., co-working on the \textit{overlap area}. The collaboration between two hands can be either asynchronous or synchronous --- An asynchronous type of control involves one hand constructing pre-conditions for the other, while a synchronous control indicates a, usually precise, mutual dependency between them.
\end{itemize}

This allows efficient control of both hands: in \textit{left area} and \textit{right area} the two hands can be controlled for different manipulations in parallel, while in \textit{overlapping area} the temporal order, i.e. synchronous or asynchronous, is made explicit depending on the task requirements and spatial relationships between objects.

\subsection{LLM Orchestration}
\begin{figure}[tb]
    \vspace*{2mm}
    \includegraphics[width=1\linewidth]{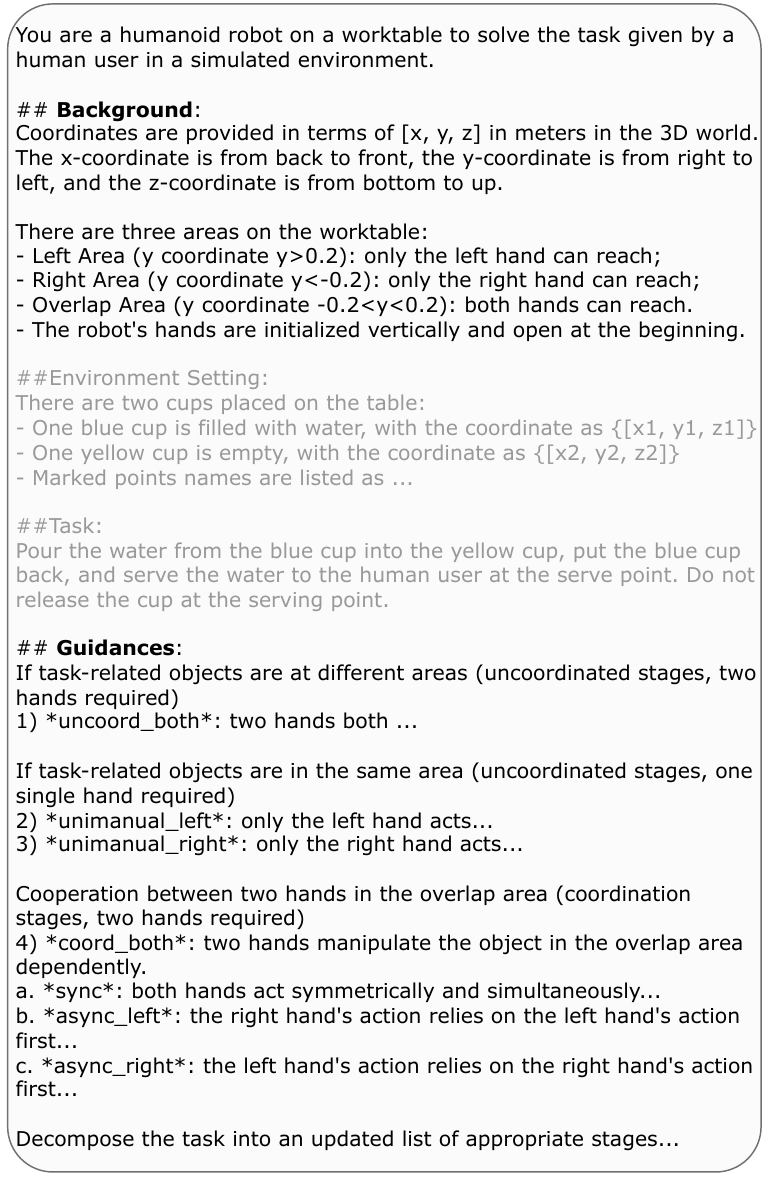}
    \caption{Prompts for the LLM to orchestrate a bimanual robot to accomplish the task. The general prompting template, indicated in standard black text, remains consistent across all tasks, while task-specific details such as environment setting and task description are shown in gray.}
    \label{fig:prompts}
\end{figure}

\begin{figure*}[!t]
\centering
\includegraphics[width=\textwidth]{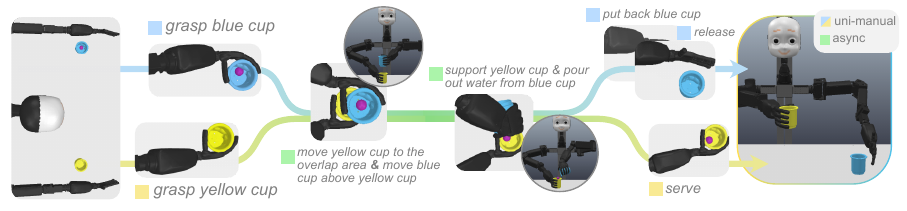}
\caption{Example of a LABOR agent's reasoning. With the guiding prompt, the LLM decomposes the task into multiple stages, i.e., uncoordinated and coordinated stages, and then generates action plans with skill primitives for both hands until the task is accomplished.}
\label{fig:preview_labor}
\end{figure*}

LLMs exhibit remarkable abilities in terms of step-by-step reasoning and instruction-following responses in robotics, which makes it possible to generate policies under temporal and spatial constraints in bimanual manipulations. As depicted in Fig. \ref{fig:prompts}, the designed prompt contains background information regarding the bimanual robot, like coordinates and workspaces of two hands, following some rules specifying the bimanual control principles in different stages. At the same time, the skill library is available to the LLM in the form of tool APIs. Given this, the prompts explicitly require the LLM to decompose the task into uncoordinated and coordinated stages and generate specific action plans based on the provided skill library. After executing an action assigned by the command, the LLM obtains observations from the environment. Iterating in this way, the LLM progressively generates a solution to the task while self-correcting mistaken commands, thus generating a chain of skills for different stages throughout the lifespan of the task. An example of LABOR agent reasoning is shown in Fig. \ref{fig:preview_labor}, where the LLM coordinator generates commands for two hands respectively based on the task requirements and the spatial relationships of the objects, and then the bimanual robot executor performs the corresponding actions in the task environment. In summary, the LABOR agent uses the LLM to explicitly coordinate two hands based on the decomposed stages while continuously generating skill-based commands, to accomplish complex long-horizon bimanual tasks effectively. 



\section{Experiments}
To determine the extent to which an LLM can manage to control the bimanual robot in everyday tasks, we deploy the LABOR agent on the semi-humanoid NICOL\footnote{The NICOL robot website can be found at https://www.inf.uni-hamburg.de/en/inst/ab/wtm/research/neurobotics/nicol.html}, the Neuro-Inspired COLlaborator \cite{Matthias2023}, on two classes of bimanual tasks in the CoppeliaSim\footnote{https://www.coppeliarobotics.com/} simulator. To validate the effectiveness of the LABOR agent in coordinating the bimanual control, we also introduce the vanilla LLM Agent as a baseline, in which our designed guiding prompts are not provided. 

\subsection{LABOR Agent Setup}
We design a skill library\footnote{The skills are developed based on the nicol\_coppeliasim Python library, https://github.com/knowledgetechnologyuhh/nicol\_coppeliasim} for the NICOL robot, mainly including two kinds of hand-crafted skills, \textit{Manipulation} skills denote the specific action designed for one hand to perform, while \textit{Information} skills are used to obtain information about the robot and the environment. Specifically, the skills with parameters shown in parenthesis are listed below.

\textit{Manipulation} skills:
\begin{itemize}
    \item \texttt{move\_to(side, obj\_name)} moves the hand on the specified \texttt{side} to the position of the object named \texttt{obj\_name}, along with a currently grasped object if any;
    \item \texttt{move\_and\_grasp(side, obj\_name)} moves the hand on the specified \texttt{side} to the position of the object named \texttt{obj\_name} and grasp it;
    \item \texttt{move\_above(side, obj\_name}) moves the hand on the specified \texttt{side} above the object named \texttt{obj\_name}, with any possibly grasped objects;
    \item \texttt{push\_to(side, source\_obj, target\_obj)} moves the hand on the specified \texttt{side} to push the object named \texttt{source\_obj} from its current position to the position of the object named \texttt{target\_obj};
    \item \texttt{pour\_out(side)} controls the hand on the specified \texttt{side} to turn the wrist to flip down. If there is an object grasped, then its content will be poured out;
    \item \texttt{release(side)} is used to open one hand on the specified \texttt{side} to release grasped objects;
    \item \texttt{reset(side)} is used to reset the hand on the specified \texttt{side} to its initial configurations;
    \item \texttt{wait(side)} is used to hold the hand on the specified \texttt{side}, including any possibly grasped objects, in its present state.
\end{itemize}

\textit{Information} skills:
\begin{itemize}
    \item \texttt{get\_arm\_state(side)} gets the status (i.e., hand positions, palm angles, and if the fingers are open or closed) of the hand on the specified \texttt{side};
    \item \texttt{get\_obj\_position(obj\_name)} gets the position of the specified \texttt{obj\_name};
\end{itemize}

These skills are designed to be generic and versatile, enabling the humanoid robot to perform complex, long-horizon manipulations involving interactions between objects and its hands. More importantly, the skills are tailored to the specific characteristics of humanoid robots, including hand movement, palm angles, and finger movements. This also points to a direction for skill design in humanoid robots: Firstly, overall hand movement toward the destination, and then, more fine-grained maneuvers provided by palm angles and finger movements. At the same time, to avoid some potential errors or collisions, some automated precondition detections are also considered in the design of the skill, for instance, when grasping a certain object, the fingers must be open, etc. In this sense, spatial dependencies or constraints within humanoid robots are established through some automatic checks, while providing flexible choices to the LLM. 

We use the OpenAI GPT-4o model as the LLM for inference in the LABOR agent. Since the accomplishment of bimanual tasks requires the LLM to generate an action plan for two hands several times, we take the LangChain\footnote{https://github.com/langchain-ai/langchain} Python library framework to enable the LLM to generate a chain of skills step by step. Based on the LangChain tool design structure, two types of tools APIs are designed: (1) \texttt{bimanual\_control(left\_command, left\_para, right\_command, right\_para)}, in which the available choices are taken from the \textit{Manipulation} skills with appropriate inputs; (2) \texttt{get\_information} functions for the \textit{Information} skills. It should be noted that the \texttt{wait} command is used to control the temporal order between two hands in asynchronous control, and when two hands are assigned the same action for the same object, then two hands will act synchronously for cooperation.

\subsection{Task Environment Setup}
\begin{figure}[htb]
    \centering
    \subfigure[real world]{
        \includegraphics[width=.47\linewidth]{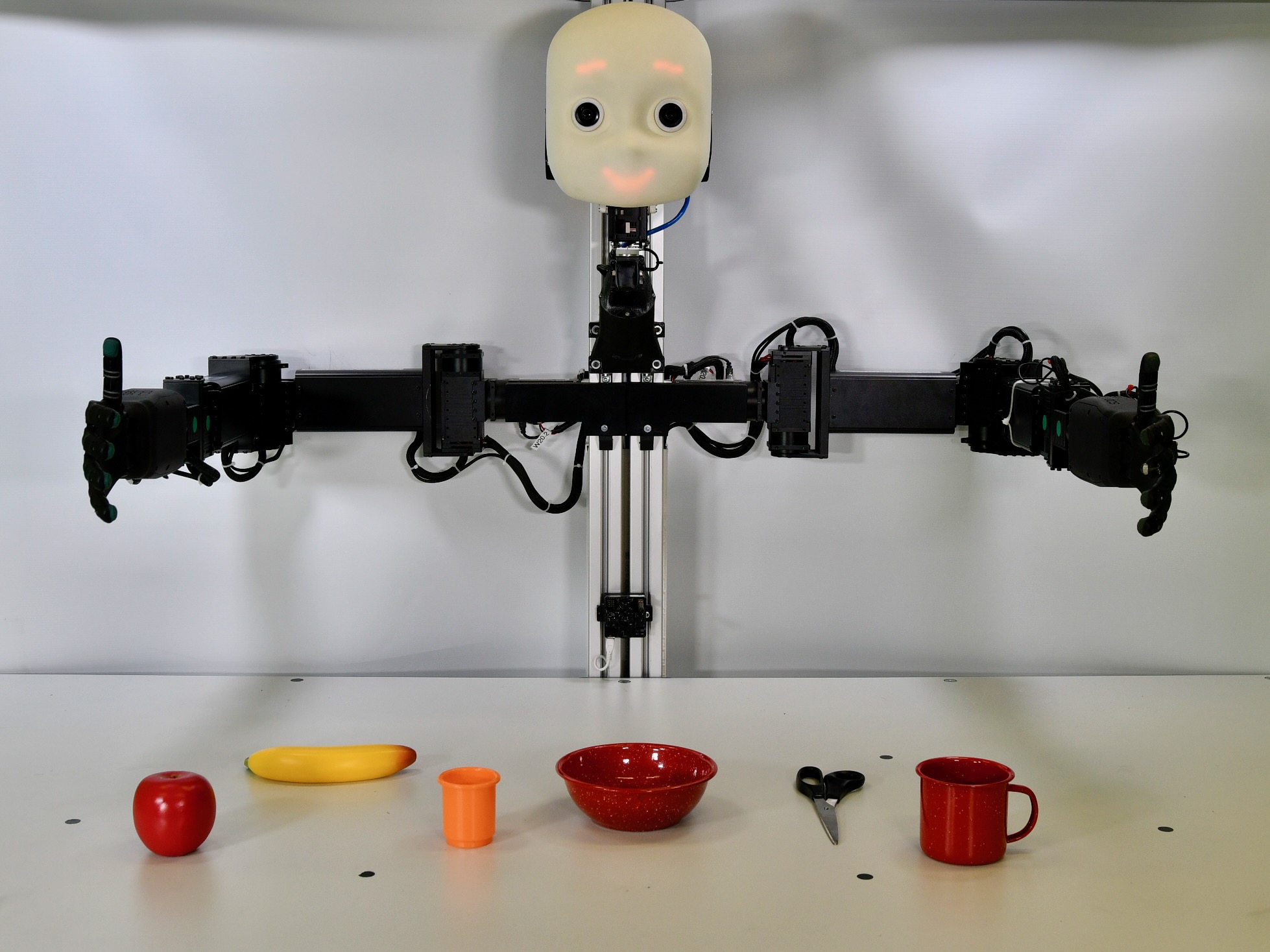}}
    \subfigure[simulated world]{
        \includegraphics[width=.47\linewidth]{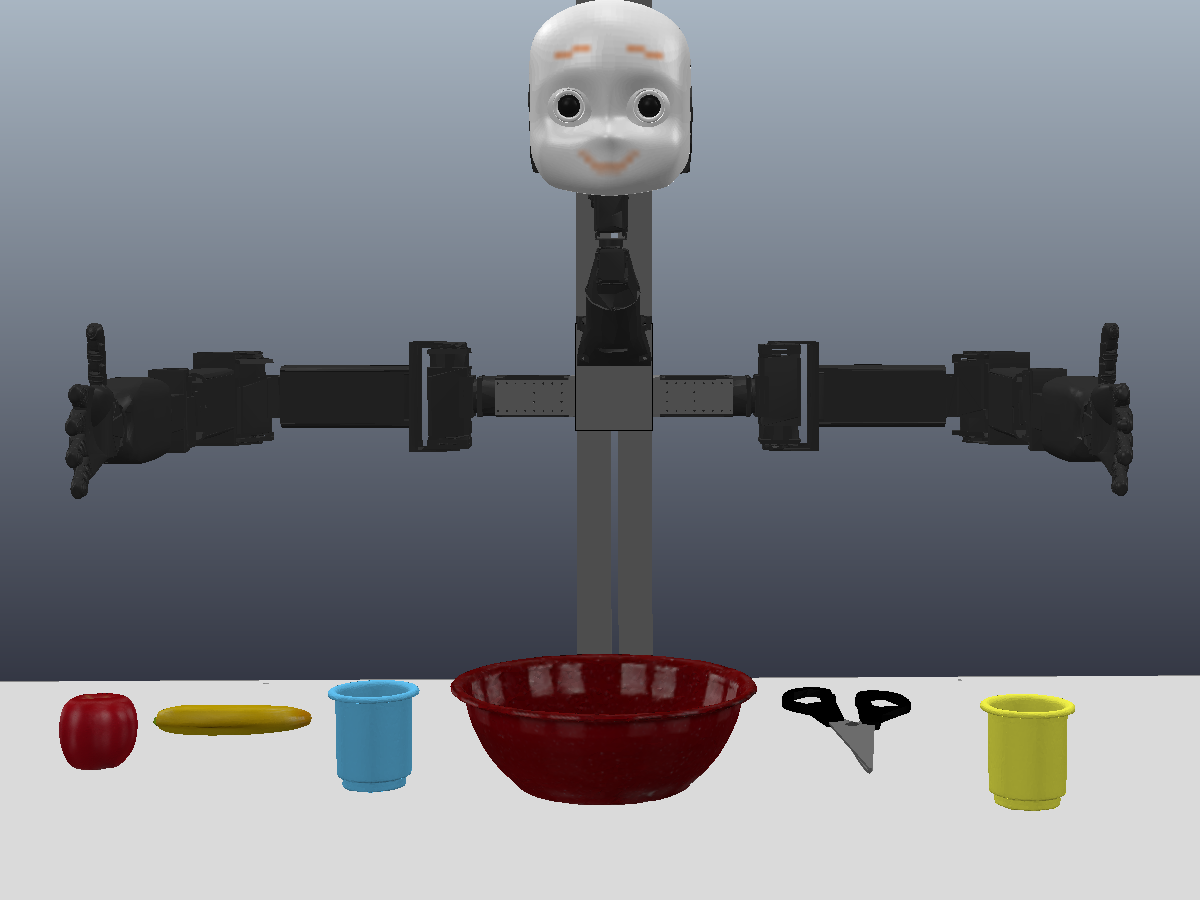}}
    \caption{NICOL workspace with daily objects in real and simulated worlds, from left to right: apple, banana, cup (orange or blue), bowl, scissors, and cup (red or yellow). In this work, experiments are designed and completed in the simulated environment, leaving real-world exploration for the foreseeable future.}
    \label{fig:nicol_workspace}
\end{figure}

NICOL has a head that can express stylized facial expressions, and two robotic \textit{Open-Manipulator-Pro} arms with adult-sized five-fingered Seed Robotics RH8D hands (for details see \cite{Matthias2023}). With such equipment, NICOL has a large bimanual workspace to handle everyday objects and tasks. NICOL's workspace with objects in the real and simulated world is shown in Fig \ref{fig:nicol_workspace}. There are two classes of tasks designed in the experiments: ServeWater and ServeFruit, which all require the cooperation of two hands with a chain of skills to accomplish. 

In the ServeWater task class, there are two cups on the work table. The yellow cup is empty and the blue cup is water-filled (we use a small ball as simulated water). The goal of the task is to use the yellow cup to serve water at a specified serving position. This means that the water has to be transferred from the blue cup to the yellow cup, which is then lifted to the serving position, while the blue cup is released back to its original position. The cups can be in two different spatial relations: in the same (both right or both left) or in different areas (one on the left and another on the right). Since the robot cannot reach with its right hand over the \textit{left area}, and vice versa, this task would require two hands to work in cooperation in some cases. However, this relationship information is not passed explicitly to the LLM, which instead receives raw information about the coordinates of the two cups, placed on the table at random positions. The task itself requires a long skill chain to accomplish and coupled with the changing positional relationship between cups and hands, choosing appropriate control policies effectively is even more challenging.

\begin{figure}[tb!]
    \centering
    \subfigure[ServeWater]{
        \includegraphics[width=0.472\linewidth]{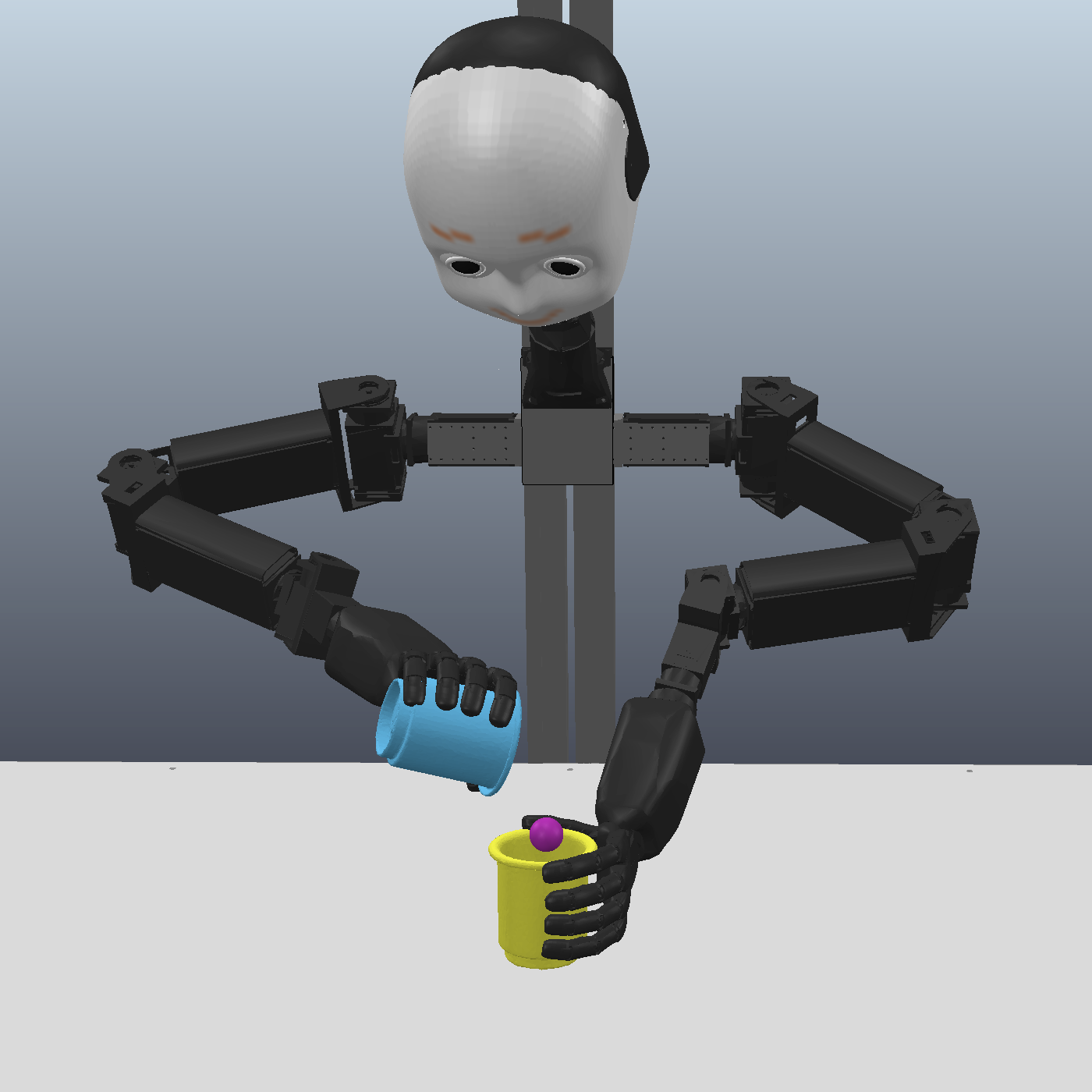}}
    \hfill
    \subfigure[ServeFruit]{
        \includegraphics[width=0.472\linewidth]{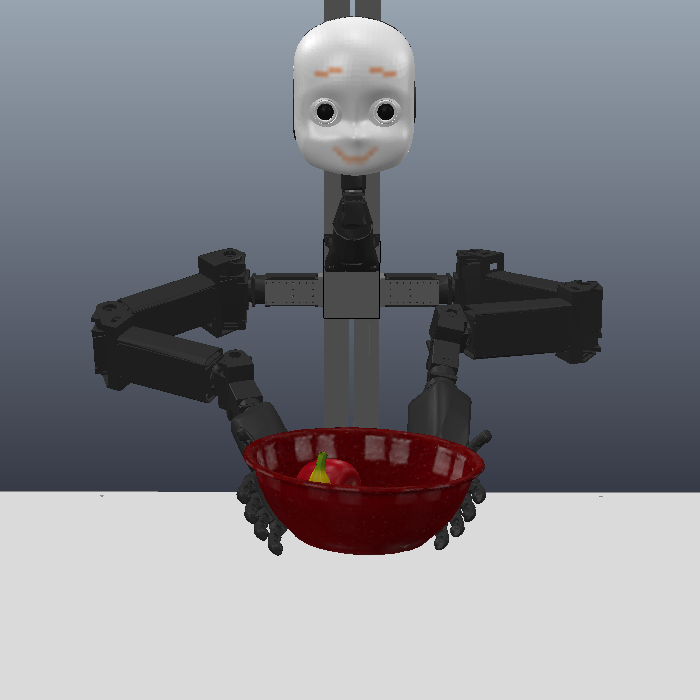} }
    \caption{Critical stages illustrated in the overlap area for two classes of tasks with the NICOL robot: (a) the right hand flips down the water-filled cup to pour out the `water' (small ball) to the empty cup while the left hand is holding the grasped empty cup; (b) left and right hands together are holding the bowl with the apple and banana inside.}
    \label{fig:nicol_tasks}
\end{figure}

In the ServeFruit task class, there is one large bowl, one apple, and one banana on the work table. The goal of the task is to grasp the fruits, release them to the bowl, and serve the bowl to the human user at the serving point. Similar to the ServeWater task, the positions of the apple, banana, and bowl bring complexity and variety to this task, i.e., the fruits are in the same or different area while the bowl can be at the left or right area, as well as the decomposition of tasks involving uncoordinated and coordinated stages. To be more specific, the grasping and lifting of the objects are independent and uncoordinated, but holding the bowl depends on two hands to hold synchronously. The most complex components also lie in that the bowl is not at the overlap area at first, thus requiring it to be pushed to the overlap area, to receive the objects from another area and be controlled by two hands.

\subsection{Results}
\begin{figure}[ht!]
    \renewcommand{\@captype}{table}
    \caption{Experiment results on two classes of bimanual robotic tasks showcase that our method outperforms baseline in terms of success rate.}
    \label{tab:success_rate}
 \begin{center}
  \centerline{\includegraphics[width=1\columnwidth]{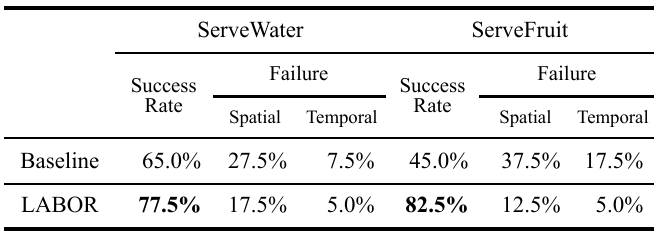}}
 \end{center}
 \vskip -0.3in
\end{figure}

In this section, we report the results of the LABOR agent in controlling the NICOL robot to solve the above tasks, including success rate and failure analysis. To evaluate the effectiveness, we repeated 40 tasks for each class, and for each time a variant of the task was randomly chosen. The success rates and failure rates in three aspects of LABOR when using GPT-4o as the LLM on these tasks are listed in Table~\ref{tab:success_rate}. It can be observed that, compared with the baseline method, the LABOR agent demonstrates higher performance in orchestrating bimanual control in long-horizon manipulation tasks, especially in the ServeFruit task. 

Regarding the properties of bimanual manipulation and the LLM itself, we found that the failure cases mainly come from two aspects namely \textit{Temporal Coordination} and \textit{Spatial Coordination}. In this sense, \textit{Temporal} failure refers to the fact that the LLM fails to generate skills in the correct order or neglects some processes during the task execution, while \textit{Spatial} failure refers that the LLM mishandles the spatial relationships in the operation, such as the relationship between two objects or between the robot's hand and an object. 

By checking the plans generated by LLMs in each failed task, we found that in the ServeWater task, the \textit{Temporal} failures mainly occurred when the pouring water process was not executed during the task accomplishment, while \textit{Spatial} failures mainly on that such process happened without moving above the yellow cup, causing the water not being transferred to another cup. Similarly in the ServeFruit task, \textit{Spatial} failures mainly come from the incorrect releasing process of the apple or banana, while the \textit{Temporal} failures are mainly due to the lack of such correct processes, causing the bowl to be served to the target point without fruits or only one single fruit. 

\makeatletter
\makeatother

In addition to this, we also see a few cases in Baseline agent where the LLM tries to perform a transfer of an object through two hands through the \textit{overlap area}, instead of performing a manipulation about two objects through the \textit{overlap area}. For example, in the Servewater task, the LLM tries to move the blue cup to another area to pour the water and then transfer it back through the overlap area, which is an interesting phenomenon, but unfortunately due to the long number of steps required for this kind of reasoning, the LLM interrupts prematurely resulting in the task not proceeding to the last step. In contrast, for the LABOR agent, the LLM innovatively uses the skill of \texttt{push\_to} to push the blue cup to the overlap area, which is feasible, but then the LLM ignores the fact that the cup is not held by the hand (requiring \texttt{move\_and\_grasp} process), failing in the task eventually.

Excluding these failure cases, one can easily observe that the LABOR agent exceeds the success rate of the Baseline agent in both long-horizon bimanual tasks, especially the ServeFruit task, demonstrating its outstanding efficacy in coordinating two hands in complex tasks.

\subsection{Discussion}

Although the guiding prompt is structured to coordinate two-hand control while allowing each hand to operate independently, it inadvertently biases the LLM towards using both hands as much as possible. Due to the generic skills we designed for the NICOL robot and the flexibility provided by the overlap area, the LLM agents occasionally exhibit unexpected behaviors. For instance, when two cups are in different areas, the most efficient approach would be to lift both cups to the overlap area to pour out the water and prepare for future actions. However, the LLM agent might instead transfer the object from the left hand to the right across the overlap area, pouring out the water in a different area.

Experimental results show that these behaviors still achieve the desired outcomes. For example, when fruits are located in different areas, the most efficient method is to lift and release the fruit on the same side as the bowl and push the bowl to the overlap area to add another fruit. However, the LLM occasionally uses the overlap area as a transition point to transfer the fruit, which, although accomplishing the task, slows down progress and incurs a higher likelihood of task failure. While we anticipated some anomalies when designing the skills, these unanticipated behaviors provide valuable insights for developing more complex skills in future real-world applications. For example, designing fine-grained precondition detection mechanisms for each skill or describing these conditions accurately in prompts so that LLMs can better understand the skill.

\section{Conclusions}

In this work, we introduce LABOR, an agent that utilizes an LLM to effectively orchestrate bimanual control when performing humanoid manipulation tasks. Under the spatio-temporal control patterns designed in guiding prompts, the LABOR agent decomposes the task into several stages involving uncoordinated and coordinated steps, according to the spatial relationships and task requirement. Based on the generated plan, the LLM then iteratively generates action steps based on the environmental feedback during task executions. Experimental results on the NICOL robot for two classes of long-horizon everyday tasks showcase the superior performance of the LABOR agent compared to the Baseline agent without guiding prompts.

\textbf{Limitations and Future Work.} Similar to other LLM-based methods, the LABOR agent is dependent on the performance of the LLM like the GPT-4o model, requiring its reasoning and understanding of the correlations of primitive skills and task solutions. Besides, since the agent is currently solely running in the simulated environment instead of the real world, some potential differences in environmental dynamics might not be sufficiently analyzed or considered, like self-collision detection between the hands, which brings some unknowns in transferring the agent. In future work, the LLM can be further used to guide the learning of versatile skills in controlling humanoid hands \cite{Ma23EurekaHumanLevel, zhao2024agentic}, which can endow the LABOR agent with fine-grained manipulation skills for everyday objects with various shapes. Moreover, vision-based foundation models \cite{yang2023set, oquab2024dinov} can enhance the scene understanding capabilities of the LABOR agent when dealing with real-world tasks. As skills expand, the LABOR agent should be able to control bimanual robots to perform more complex long-horizon tasks.

\section{Acknowledgment}
The authors would like to thank OpenAI and their Researcher Access Program for generously providing GPT-4o API tokens support, and Jan-Gerrit Habekost and Lennart Clasmeier for their work and support for the NICOL robot in the simulation.

\addtolength{\textheight}{-5.5cm}



\bibliographystyle{IEEEtran}
\bibliography{references, llmrobot}
\end{document}